\documentclass[11pt]{article}

\usepackage[final]{acl}

\usepackage{times}
\usepackage{latexsym}
\usepackage{amsmath}

\usepackage[T1]{fontenc}

\usepackage[utf8]{inputenc}

\usepackage{microtype}

\usepackage{inconsolata}

\usepackage{graphicx}

\title{Evaluating LLMs’ Reasoning Over Ordered Procedural Steps}

\author{Adrita Anika\textsuperscript{1}\thanks{Work done outside of role at Amazon}, Md Messal Monem Miah\textsuperscript{2}\\ 
\textsuperscript{1}Amazon\\
\textsuperscript{2}Texas A\&M University\\
\normalsize{\texttt{adritaanika23@gmail.com}},
\normalsize{\texttt{messal.monem@tamu.edu}}}

\begin{document}
\maketitle
\begin{abstract}

Reasoning over procedural sequences, where the order of steps directly impacts outcomes, is a critical capability for large language models (LLMs). In this work, we study the task of reconstructing globally ordered sequences from shuffled procedural steps, using a curated dataset of food recipes, a domain where correct sequencing is essential for task success. We evaluate several LLMs under zero-shot and few-shot settings and present a comprehensive evaluation framework that adapts established metrics from ranking and sequence alignment. These include Kendall’s Tau, Normalized Longest Common Subsequence (NLCS), and Normalized Edit Distance (NED), which capture complementary aspects of ordering quality. Our analysis shows that model performance declines with increasing sequence length, reflecting the added complexity of longer procedures. We also find that greater step displacement in the input, corresponding to more severe shuffling, leads to further degradation. These findings highlight the limitations of current LLMs in procedural reasoning, especially with longer and more disordered inputs.


\end{abstract}

\section{Introduction}

Understanding and generating correctly ordered action sequences is a key aspect of reasoning. Many real world tasks, such as cooking recipes or carrying out technical procedures, require steps be completed in a precise order to achieve the intended outcome. LLMs have demonstrated strong performance on various reasoning tasks including arithmetic computation \citep{imani2023mathprompter, ahn2024large}, commonsense inference \citep{rajani2019explain}, question answering \citep{robinson2022leveraging, anika2025leveraging}, and multimodal reasoning and understanding tasks \citep{miah-etal-2025-hidden, miah-etal-2023-hierarchical}. While much prior work has evaluated LLMs on step-by-step reasoning, their ability to reason over and reconstruct ordered procedural steps remains relatively underexplored.

\begin{figure}[t!]
\includegraphics[width=7.4cm]{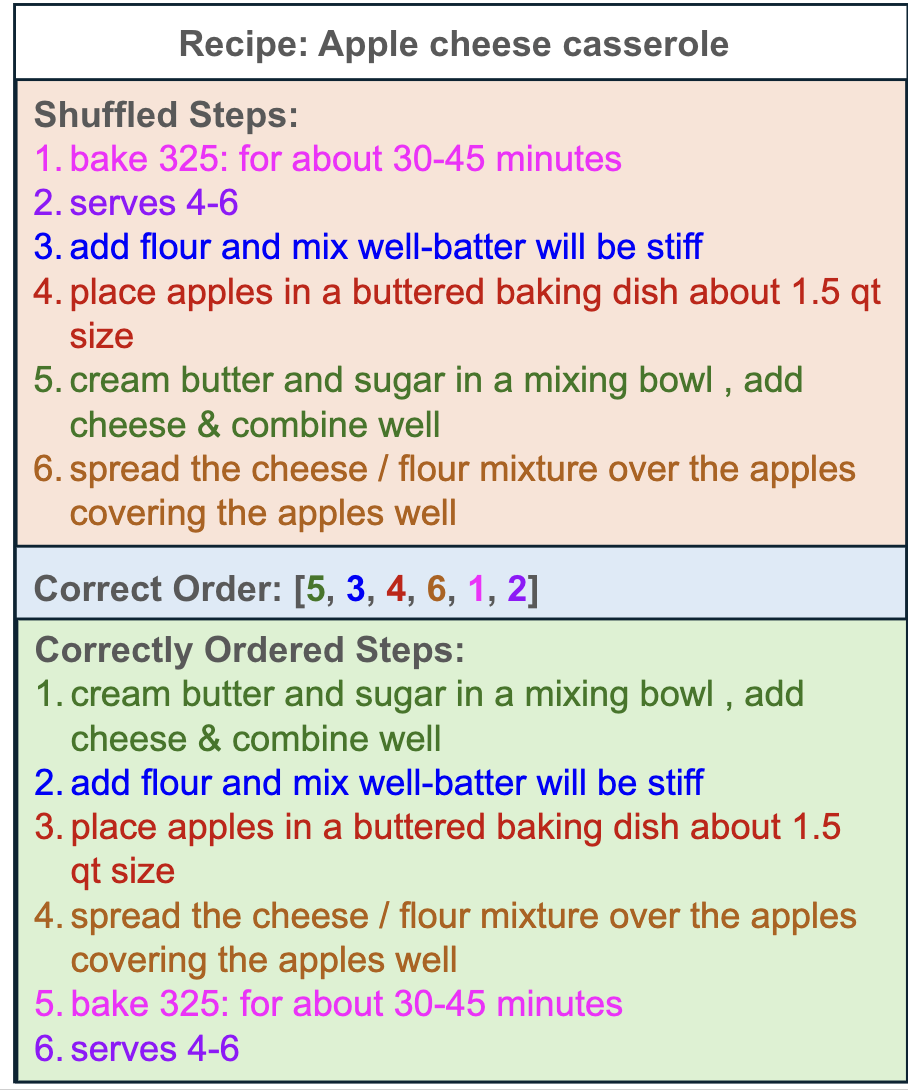}
\caption{Example of the step ordering task. Given a shuffled list of recipe instructions (top), the goal is to recover the correct sequence (bottom) required to successfully complete the recipe. The middle row shows the gold permutation that reorders the input into the correct order.}
\label{task}
\end{figure}

Step ordering tasks, where the correctness of the output depends on recovering a globally coherent sequence, pose a unique challenge. Most existing research focuses on predicting the immediate next step \citep{yong2025attackseqbench, wang2023steps}, rather than reconstructing the full sequence from a shuffled set. Moreover, prior evaluations rely only on accuracy \citep{quan2024econlogicqa}, measuring exact matches between predicted and reference positions. This limits our ability to fully understand LLMs procedural reasoning. In this work, we evaluate LLMs' step ordering capabilities using a curated dataset of food recipes due to their clearly defined structure and strong ordering constraints. As illustrated in Figure~\ref{task}, the model receives a shuffled list of recipe instructions and must recover the correct sequence that reflects the intended preparation process.  We use complementary metrics, including Kendall’s Tau to measure rank correlation, Normalized Longest Common Subsequence (NLCS) to assess subsequence preservation, and Normalized Edit Distance (NED) to quantify reordering cost, providing a deeper analysis of model performance. We conduct a systematic evaluation across multiple LLMs under 0-shot and few-shot settings. We further analyze performance as a function of sequence complexity, examining how models respond to longer recipes and greater amounts of step shuffling. Our main contributions are:



\begin{itemize}
    \item We present a comprehensive evaluation of step-order reasoning abilities in LLMs using a structured cooking recipe dataset, going beyond next-step prediction to assess full sequence reconstruction.
    \item We introduce a multi-metric evaluation framework that captures partial correctness, subsequence alignment, and reordering cost—offering a richer picture of model behavior than accuracy alone.
    \item We analyze how model performance varies with step count and shuffling difficulty, revealing performance gaps and highlighting ongoing challenges in LLM's procedural reasoning
\end{itemize}

\section{Related Work}

Previous studies have explored LLMs reasoning on procedural tasks. STEPS \cite{wang2023steps} proposes a benchmark to assess models' procedural reasoning through two subtasks: next-step prediction and multiple-choice selection of the correct next step. While valuable, these tasks focus only on local coherence by predicting or identifying a single correct step rather than requiring the model to recover an entire global sequence. ProcBench \cite{fujisawa2024procbench} focuses on multi-step reasoning over structured tasks like string manipulation and arithmetic operations. It evaluates whether LLMs can follow explicit instructions step-by-step, minimizing the need for external knowledge or path exploration. AttackSeqBench \cite{yong2025attackseqbench} evaluates LLMs’ understanding of sequential patterns in cybersecurity reports through a suite of question-answering tasks. These are designed to probe models’ ability to reason about adversarial behavior over time. However, the setting remains extractive QA, and models are not required to reconstruct full procedural chains. EconLogicQA \cite{quan2024econlogicqa} introduces a benchmark targeting sequential reasoning over interdependent events drawn from economic articles, emphasizing complex temporal and logical relationships. However, like other QA-style evaluations, it relies mainly on accuracy or exact match at each step, missing partial correctness or structural misalignment. In contrast, our study focuses on full-sequence reconstruction and introduces additional metrics for a more comprehensive assessment of procedural reasoning.


\section{Problem Definition}
Given a shuffled set of procedural steps \( S = \{s_1, s_2, \ldots, s_n\} \), the goal is to find a permutation \(\hat{S} = \{\hat{s}_1, \hat{s}_2, \ldots, \hat{s}_n\} \) that best approximates the ground truth ordered sequence \( S^* = \{s^*_1, s^*_2, \ldots, s^*_n\} \). The predicted sequence \(\hat{S}\) is aligned with \(S^*\) to assess ordering quality.

\begin{figure}[t!]
    \includegraphics[width=7.3cm]{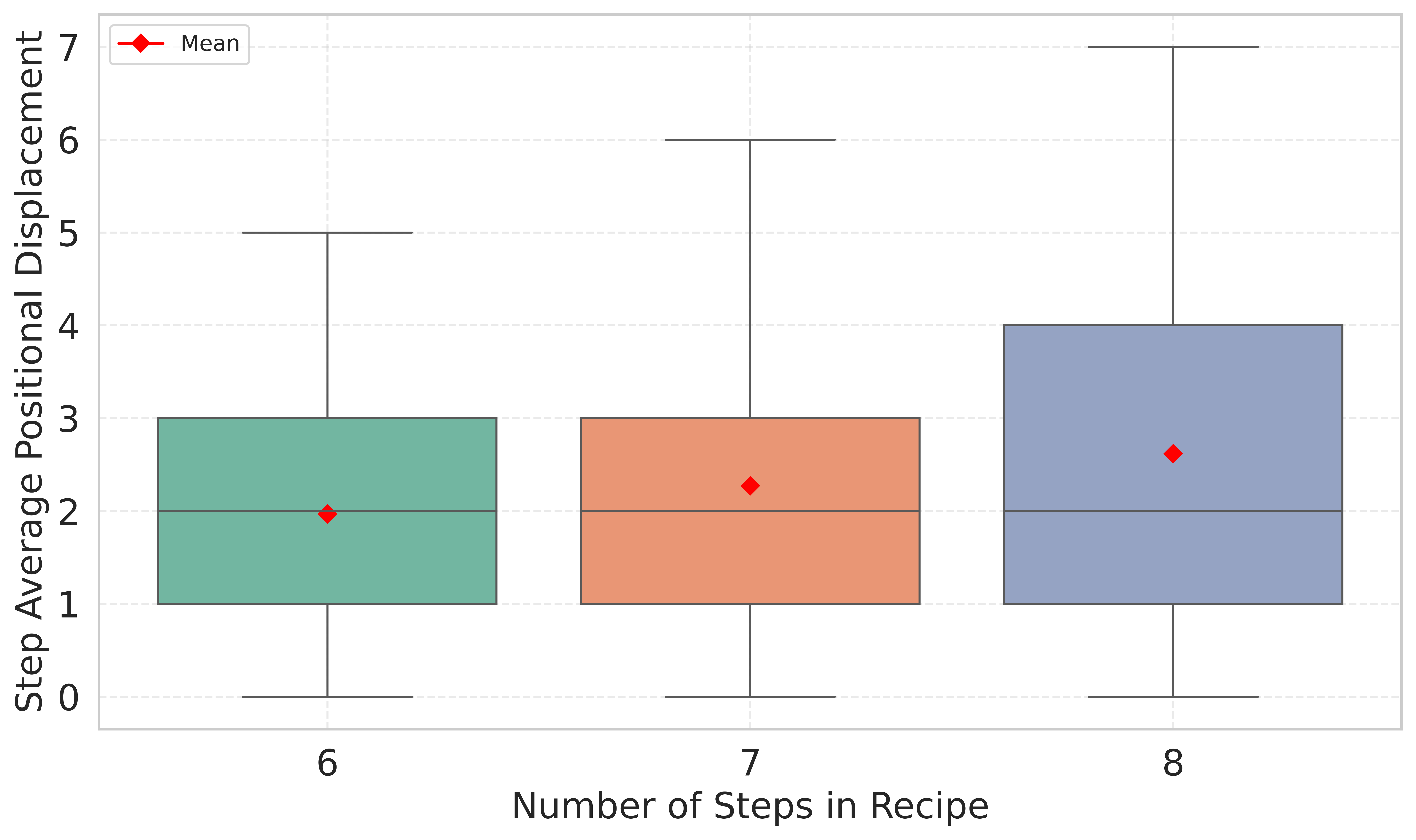}
    \caption{Distribution of step movement distances across recipes of different sequence lengths.}
    \label{fig:data_dist}
\end{figure}

\section{Dataset}
\citet{majumder2019generating} has introduced a dataset containing 230K recipes from Food.com\footnote{\url{https://www.food.com}}. From this corpus, we  select 5,000 samples with 6 to 8 steps and 5 to 6 ingredients, ensuring moderate sequence length and complexity. Food recipes are inherently sequential, and prior work has treated step ordering as critical to successful execution~\citep{wang2023steps}. However, some recipes may have some steps that may be interchangeable without affecting the outcome (e.g., cutting onions and cutting potatoes). To focus on sequences where step order is necessary, we apply an additional curation step using a LLM to filter recipes requiring strict ordering (see Appendix \ref{sec:appendix A}) which yields to 1,740 recipes. Each recipe provides a coherent step sequence \( S = \{s_1, \ldots, s_n\} \), which we shuffle randomly (with fixed seed) to produce \( \hat{S} \). The task is to recover the original order from \( \hat{S} \). We generate a permutation label \( \pi \in \{1, \ldots, n\}^n \), where \( \pi_i \) denotes the original position of the \( i \)-th step in the shuffled sequence. 

The dataset is balanced with \(29.6\%\) (515 samples) having 6 steps, \(36.7\%\) (638 samples) with 7 steps, and \(33.7\%\) (587 samples) with 8 steps. 

We quantify the extent of step permutation by measuring the \emph{average positional displacement}, defined as the mean absolute difference between the original position \( p_i \) and the shuffled position \( s_i \) of each step \( i \) in a sequence of length \( n \), i.e.,

\[
\frac{1}{n} \sum_{i=1}^n |p_i - s_i|
\]

This metric captures the average magnitude of step movement caused by shuffling. Figure~\ref{fig:data_dist} visualizes the displacement distribution, highlighting mean values and variability. Longer sequences show higher average displacement, increasing from 1.97 for 6-step to 2.62 for 8-step recipes, indicating greater complexity in step rearrangements. The median displacement remains at 2 across all lengths.

\section{Experimental Setup}

\subsection{Inference Settings}

We evaluate five instruction-tuned LLMs: \texttt{Llama-3.1-8B-Instruct}\citep{llama3.1}, \texttt{Mistral-7B-Instruct}\citep{jiang2024mistral}, \texttt{Gemma-2-9b-it}\citep{gemma2}, \texttt{GPT-4o-mini}\citep{gpt4o} and \texttt{Qwen3-8b}\citep{qwen3technicalreport} under three settings: zero-shot, 3-shot, and 5-shot. 


Each model is prompted with a set of task instructions, the recipe name, and a list of shuffled procedural steps (see Appendix \ref{sec:appendix B}, \ref{sec:appendix C} ). The model is expected to output both the reordered step sequence and the corresponding order as a list of indices. Qwen3-8b is a reasoning model, and thinking mode was enabled during evaluation. From the  1,740 samples, we use 1,700 as the test set and remaining samples for few-shot demonstrations.

\subsection{Evaluation Metrics}
We use four complementary metrics:

\paragraph{Step Accuracy (Acc).}  
We report accuracy at the step level:

\[
\text{Acc} = \frac{1}{n} \sum_{i=1}^{n} (\hat{\pi}_i = \pi_i)
\]

This metric measures the fraction of steps placed at the correct positions and provides a measure of how often the model recovers the exact step location.

\paragraph{Kendall's Tau (\(\tau\)) (KTau).}  
Kendall's tau is a rank correlation metric ~\citep{lapata2006automatic} that evaluates the relative order of all possible step pairs between the predicted permutation \(\hat{\pi}\) and the ground truth \(\pi\). It is computed as

\[
\tau = \frac{C - D}{\frac{1}{2}n(n-1)}
\]

where \(C\) is the number of concordant pairs and \(D\) is the number of discordant pairs. It is suitable for assessing whether the predicted step sequence agrees with the ground truth in terms of relative step precedence, regardless of their absolute positions. This metric is sensitive to pairwise inversions and captures global ordering consistency.

\paragraph{Normalized Edit Distance (NED).}  
Edit distance counts the number of insertions, deletions, or swaps required to convert the predicted order into the gold sequence. We use its normalized form~\citep{marzal2002computation},

\[
\text{NED} = \frac{\text{EditDistance}(\hat{\pi}, \pi)}{n}
\]

This metric measures the total transformation cost and is particularly sensitive to local misplacements. NED is an error-based metric; lower values indicate better sequence similarity.

\paragraph{Normalized Longest Common Subsequence (NLCS).}  
We compute the length of the longest common subsequence (LCS) between \(\hat{\pi}\) and \(\pi\), normalized by the length of the reference:

\[
\text{NLCS} = \frac{\text{LCS}(\hat{\pi}, \pi)}{n}
\]

This metric rewards the preservation of correct subsequences and reflects the extent to which a model recovers partial ordering structure. It is robust to small local reorderings and has been widely used in structured sequence evaluation.

\paragraph{}  
Together, these metrics capture {global, local, and partial structural alignment between predicted and target step sequences.

\begin{table}[h!]
\centering
\small
\resizebox{\columnwidth}{!}{
\begin{tabular}{|l|c|c|c|c|c|}
\hline
\textbf{Model} & \textbf{Shots} & \textbf{Acc} & \textbf{NLCS} & \textbf{KTau} & \textbf{NED} \\
\hline
Llama-3.1 & 0-shot & 0.33 & 0.62 & 0.70 & 0.56 \\
           & 3-shot & 0.45 & 0.73 & 0.83 & 0.42 \\
           & 5-shot & 0.44 & 0.73 & 0.83 & 0.43 \\
\hline
Mistral   & 0-shot & 0.29 & 0.61 & 0.73 & 0.55 \\
           & 3-shot & 0.32 & 0.66 & 0.79 & 0.51 \\
           & 5-shot & 0.31 & 0.66 & 0.79 & 0.51 \\
\hline
Gemma-2   & 0-shot & 0.59 & 0.81 & 0.87 & 0.32 \\
           & 3-shot & 0.62 & 0.84 & \underline{0.90} & 0.28 \\
           & 5-shot & 0.61 & 0.84 & \underline{0.90} & 0.28 \\
\hline
GPT-4o    & 0-shot & 0.63 & 0.83 & 0.89 & 0.29 \\
           & 3-shot & \underline{0.64} & \underline{0.85} & \underline{0.90} & \underline{0.27} \\
           & 5-shot & \underline{0.64} & 0.84 & \underline{0.90} & \underline{0.27} \\
\hline
Qwen3     & 0-shot & \textbf{0.71} & \textbf{0.88} & \textbf{0.92} & \textbf{0.22} \\
           & 3-shot & 0.63 & 0.82 & 0.88 & 0.30 \\
           & 5-shot & 0.62 & 0.81 & 0.87 & 0.30 \\
\hline
\end{tabular}
}
\caption{Performance of different models across few-shot settings (0, 3, 5) using Accuracy (Acc), Normalized Longest Common Subsequence (NLCS), Kendall Tau (KTau), and Normalized Edit Distance (NED). The best and second-best results across all models are highlighted (lowest for NED).}
\label{tab:perf}
\end{table}

\section{Results and Analysis}
\subsection{Performance in Zero-Shot and Few-Shot Settings}

Table~\ref{tab:perf} reports LLMs’ performance in 0-shot and few-shot settings. Most models, i.e., Llama-3.1, Mistral, Gemma-2, and GPT-4o, show notable improvements from 0-shot to 3-shot prompting, whereas Qwen3 maintains strong performance even without demonstrations. This suggests that a small number of demonstrations helps models learn structural reordering patterns. However, across all models, performance plateaus beyond 3-shot as no model shows meaningful gains with five examples, indicating limited additional value from further demonstrations.

Qwen3 achieves the best overall performance across all metrics in the 0-shot setting, reaching the highest accuracy (0.71), NLCS (0.88), and KTau (0.92), and the lowest NED (0.22), indicating strong intrinsic capabilities for accurate and consistent reordering than other models. This superior performance suggests that Qwen3’s reasoning abilities allow it to better understand and model the logical structure of sequences, enabling it to maintain correct absolute positions and preserve subsequences effectively. GPT-4o ranks second overall, performing best among the remaining models in the 3-shot setting, with high accuracy (0.64), NLCS (0.85), and KTau (0.90), and low NED (0.27), indicating better absolute positioning, strong preservation of subsequences, and minimal local reordering. Gemma-2 performs competitively whereas Mistral and Llama-3.1 fall behind across all metrics, often producing more fragmented sequences (lower NLCS) and higher reordering costs (higher NED), despite moderate KTau scores.

KTau values show that even when models make positional errors, they may still preserve correct relative ordering. For example, Llama-3.1 in 3-shot achieves 0.83 KTau despite only 0.45 accuracy, indicating good understanding of step precedence even with absolute misplacements. NED values further expose models’ tendency to make local misorderings, with Qwen3 achieving the lowest scores, followed by GPT-4o and Gemma-2, indicating minimal local reordering. NLCS emphasizes preservation of long subsequences; again, Qwen3 attains the highest score, demonstrating stronger retention of step continuity compared to other models. Despite these improvements, all models still exhibit gaps in fine-grained step-level reasoning, suggesting remaining challenges in capturing detailed procedural structure.

\begin{figure}[t!]
    \centering
    \includegraphics[width=7.5cm]{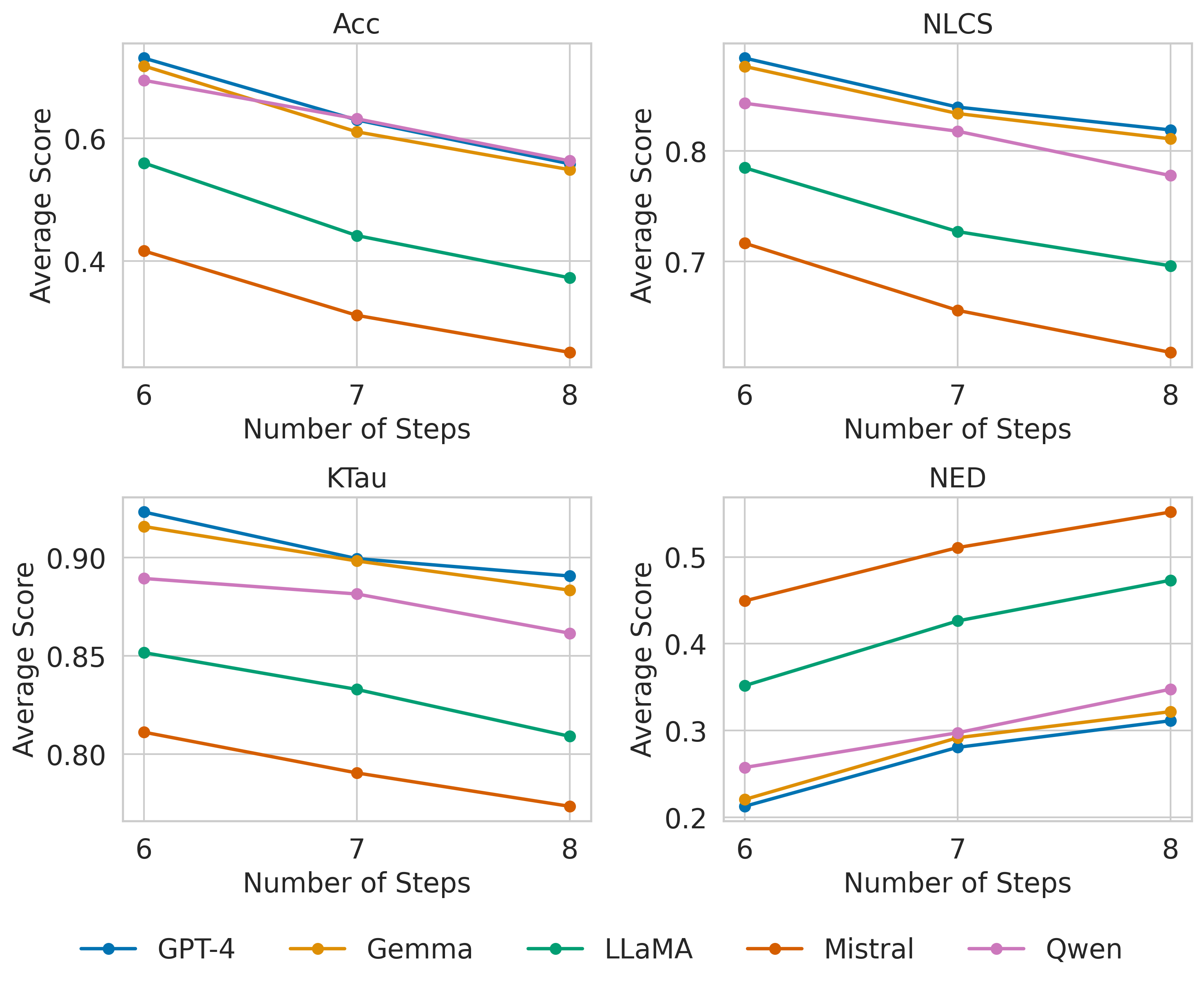}
    \caption{3-shot performance of models (Acc, NLCS, KTau, NED) across varying numbers of steps.}
    \label{fig:step_movement}
\end{figure}

\subsection{Impact of Number of Steps on Ordering Performance}
We analyze model performance in the three-shot setting by the number of steps in the sequence (\texttt{n}), where longer sequences indicate increased complexity. As shown in Figure~\ref{fig:step_movement}, with \texttt{n} increasing from 6 to 8, a general performance decline is observed across all models, reflecting the added difficulty in recovering longer step sequences. GPT-4 consistently performs best, maintaining high accuracy (0.73~$\rightarrow$~0.56), strong subsequence alignment (NLCS: 0.88~$\rightarrow$~0.82), and low edit cost (NED: 0.21~$\rightarrow$~0.31) as complexity increases. Gemma 2B shows similar robustness, with slightly lower performance. 
Qwen 3 performs competitively, surpassing LLaMA 3.1 and Mistral on most metrics, with stable subsequence alignment (NLCS $\approx$ 0.86~$\rightarrow$~0.79) and consistent ranking correlation (KTau $\approx$ 0.89~$\rightarrow$~0.86).
In contrast, LLaMA 3.1 and Mistral show more pronounced declines across all metrics, indicating a stronger tendency to produce fragmented and disordered outputs under increased complexity.

\begin{figure}[t!]
    \centering
    \includegraphics[width=7.4cm]{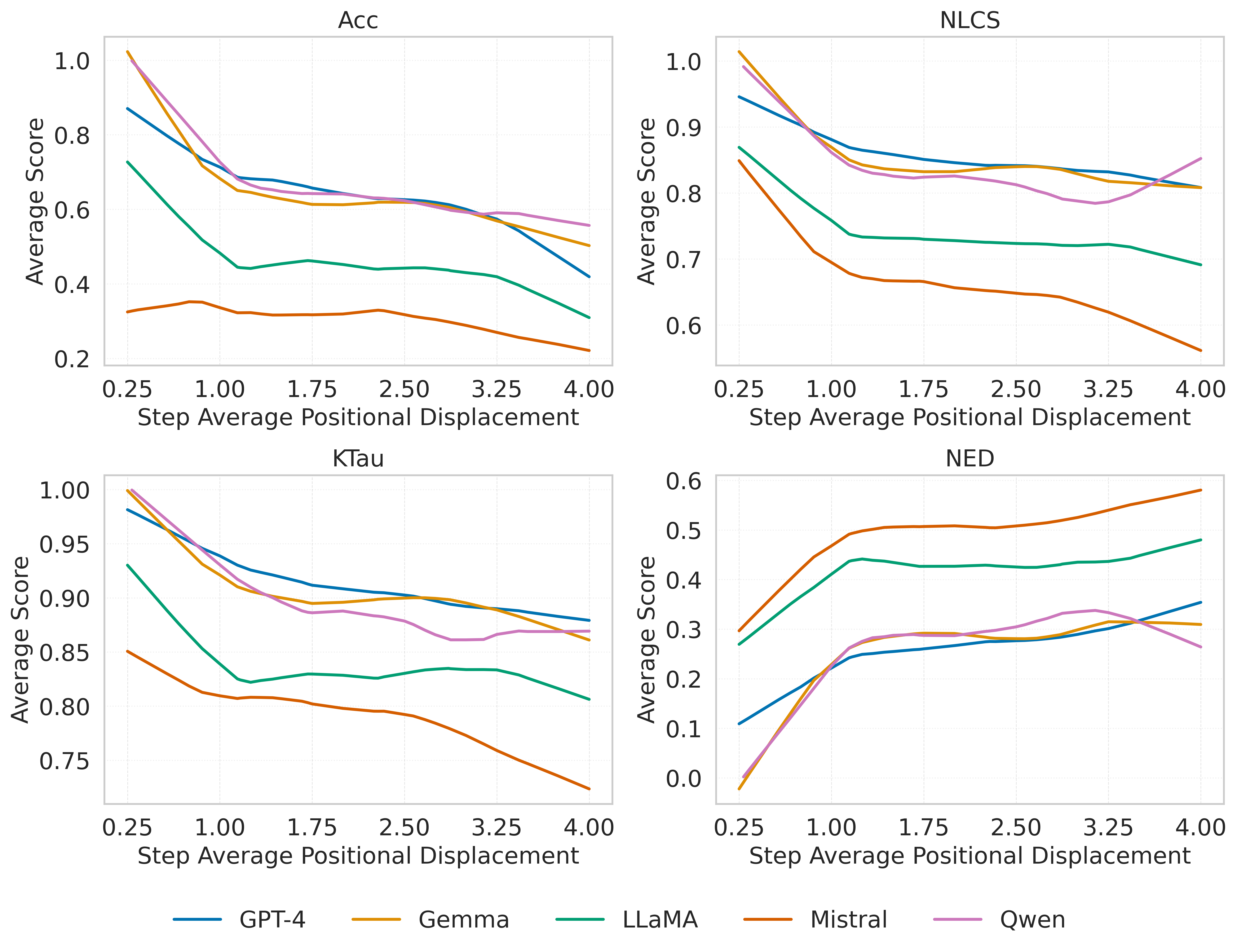}
    \caption{Smoothed 3-shot performance of models (Acc, NLCS, KTau, NED) across average positional displacement}
    \label{fig:avg_move}
\end{figure}


\subsection{Impact of Step Average Positional Displacement on Model Performance}
We further assess robustness to reordering by analyzing model performance with respect to \emph{average positional displacement}. As shown in Figure~\ref{fig:avg_move}, as displacement increases, indicating more severe shuffling, all models exhibit noticeable performance degradation across metrics. For Accuracy, Gemma-2 starts highest but drops sharply from near 1.0 to approximately 0.4, while GPT-4 declines more gradually, demonstrating greater stability. Qwen3 achieves strong initial performance and remains competitive with GPT-4, showing moderate declines in accuracy and NLCS as displacement increases. In contrast, LLaMA 3.1 remains consistently lower, and Mistral performs worst overall, maintaining a steady but low trajectory across all displacement levels. For NLCS, GPT-4, Gemma-2, and Qwen3 maintain relatively high subsequence alignment compared to LLaMA 3.1 and Mistral. KTau follows a similar trend, with GPT-4 and Qwen3 sustaining high ranking correlation and smaller declines under increased displacement. NED rises with displacement, reflecting larger deviations from the reference ordering, where GPT-4 and Qwen show smaller increases compared to the other models, indicating stronger resistance to disorder.


\section{Conclusion}
We evaluated five LLMs on step ordering tasks using four complementary metrics. Most models improved from 0-shot to 3-shot prompting, with no additional gains from five examples, whereas Qwen3 maintained strong performance in the 0-shot setting but degraded in the 3-shot setting. No gains were observed from 3-shot to 5-shot for any model. Qwen3 achieved the best overall performance, followed by GPT-4o and Gemma-2, while Llama-3.1 and Mistral were less reliable. Performance declined as sequence length and reordering complexity increased. Although models often preserved relative ordering (high KTau) and subsequences (high NLCS), they continued to struggle with precise step-level reasoning, highlighting limitations in LLMs’ procedural understanding.


\section{Limitations}
While our study offers a comprehensive evaluation of LLMs on step ordering tasks, it leaves room for further exploration. First, we restrict our analysis to relatively short sequences (6–8 steps), extending the evaluation to longer instructions could uncover new insights. Second, we evaluate only instruction-tuned models without task-specific fine-tuning. Targeted fine-tuning on step ordering or procedural datasets may yield improved performance. Finally, although our dataset is carefully curated to ensure strong ordering constraints, it is focused solely on the cooking domain; evaluating cross-domain generalization  would offer a broader view of LLM procedural reasoning capabilities.

\section{Ethics Statement}
The research conducted for this paper adheres to ethical principles and guidelines. The study utilizes publicly available datasets from reputable sources, ensuring compliance with data usage policies and respecting the privacy and confidentiality of individuals involved. All methodologies follow established scientific practices, emphasizing transparency, validity, and reliability. As the study does not involve human subjects or sensitive information, no ethics approval was sought. 

\bibliography{main.bbl}

\clearpage

\appendix

\section{Dataset Curation}
\label{sec:appendix A}
\subsection{Dataset Curation Prompt}
We used LLaMA-3 models with the prompt shown in Figure~\ref{fig:pt1} to curate a dataset of 5,000 samples. Each sample was processed in two independent runs, where the model was asked to determine whether the order of steps matters. We retained only those samples for which both runs returned a positive response (“yes”), indicating that step order is important.  

\subsection{Dataset Curation Examples}
Some examples of data curation are provided in Figure~\ref{fig:pt11}. The examples demonstrate which food recipes was selected by the LLMs along with its reasoning.

\section{Task Details}
\label{sec:appendix B}
\subsection{Prompt}
The prompt used for evaluation of the LLMs for step ordering is given in Figure~\ref{pt2}. This is the prompt for 0-shot setting. Demonstrations were incorporated for 3-shot and 5-shot settings.

\begin{figure}[ht]
\centering
\includegraphics[width=15cm]{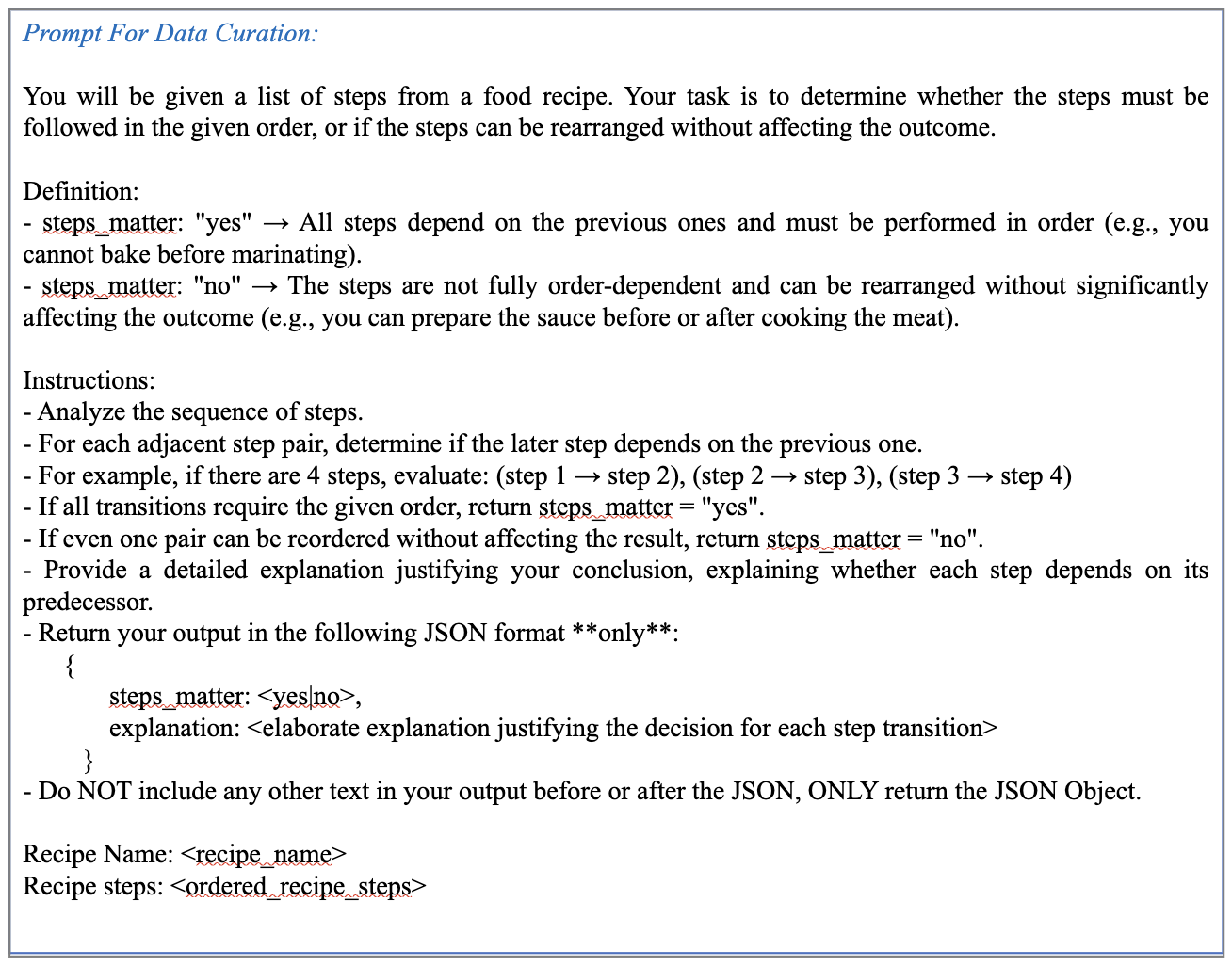}
\caption{LLM prompt used for data curation}
\label{fig:pt1}
\end{figure}

\subsection{Few-Shot Examples Selections}
The few shot examples in this experiment were chosen randomly out of the samples considered for few shot demponstartions. For 3 shot setting- 1 example with 6 steps, 1 example with 7 steps, 1 example with 8 steps were chosen.  For 5 shot setting- 2 examples with 6 steps, 1 example with 7 steps, 2 examples with 8 steps were chosen.  These fixed set of chosen examples were used for all evaluation.

\clearpage

\begin{figure}[h!]
\centering
\includegraphics[width=12cm]{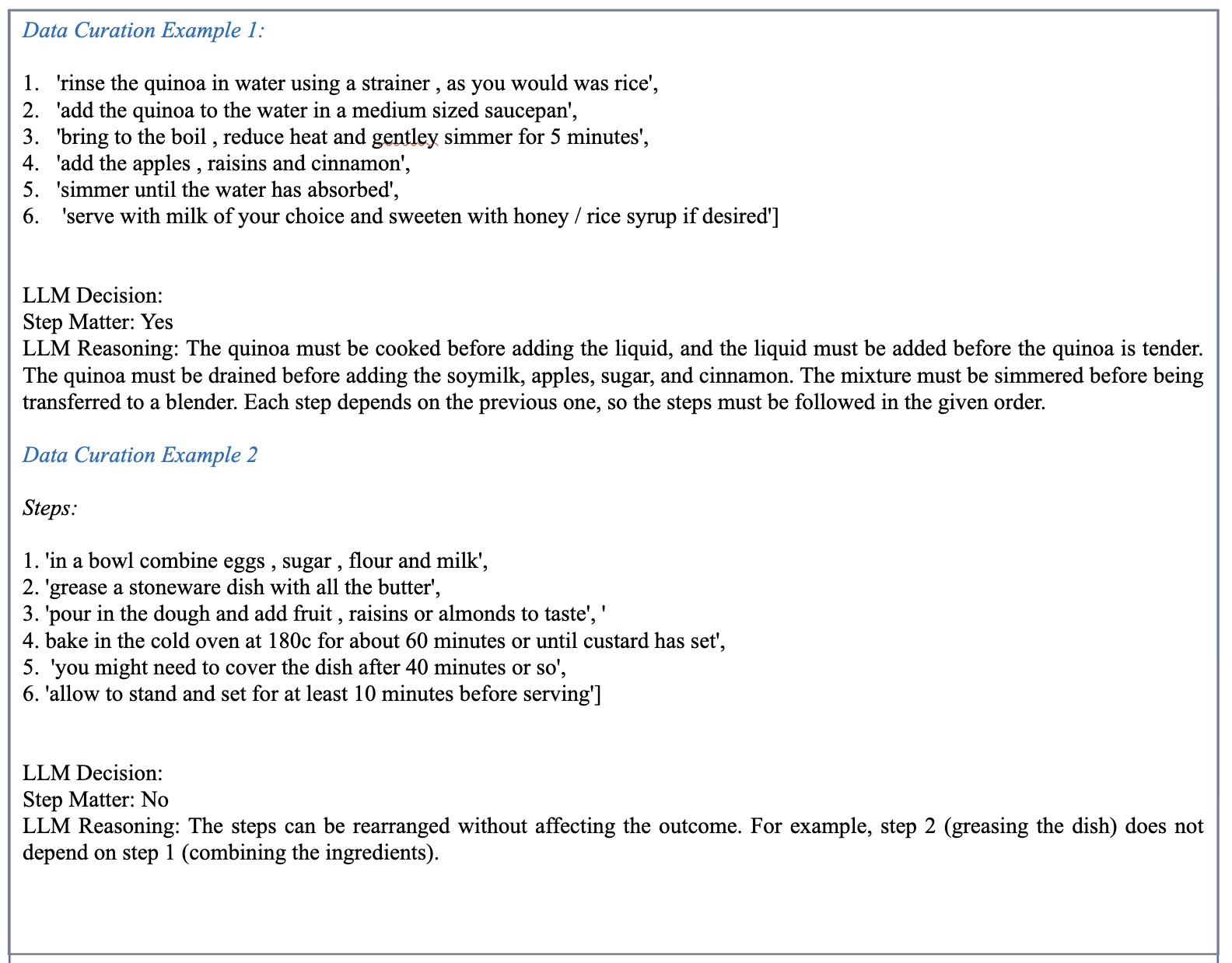}
\caption{Examples of data curation}
\label{fig:pt11}
\end{figure}

\begin{figure}[h!]
\centering
\includegraphics[width=12cm]{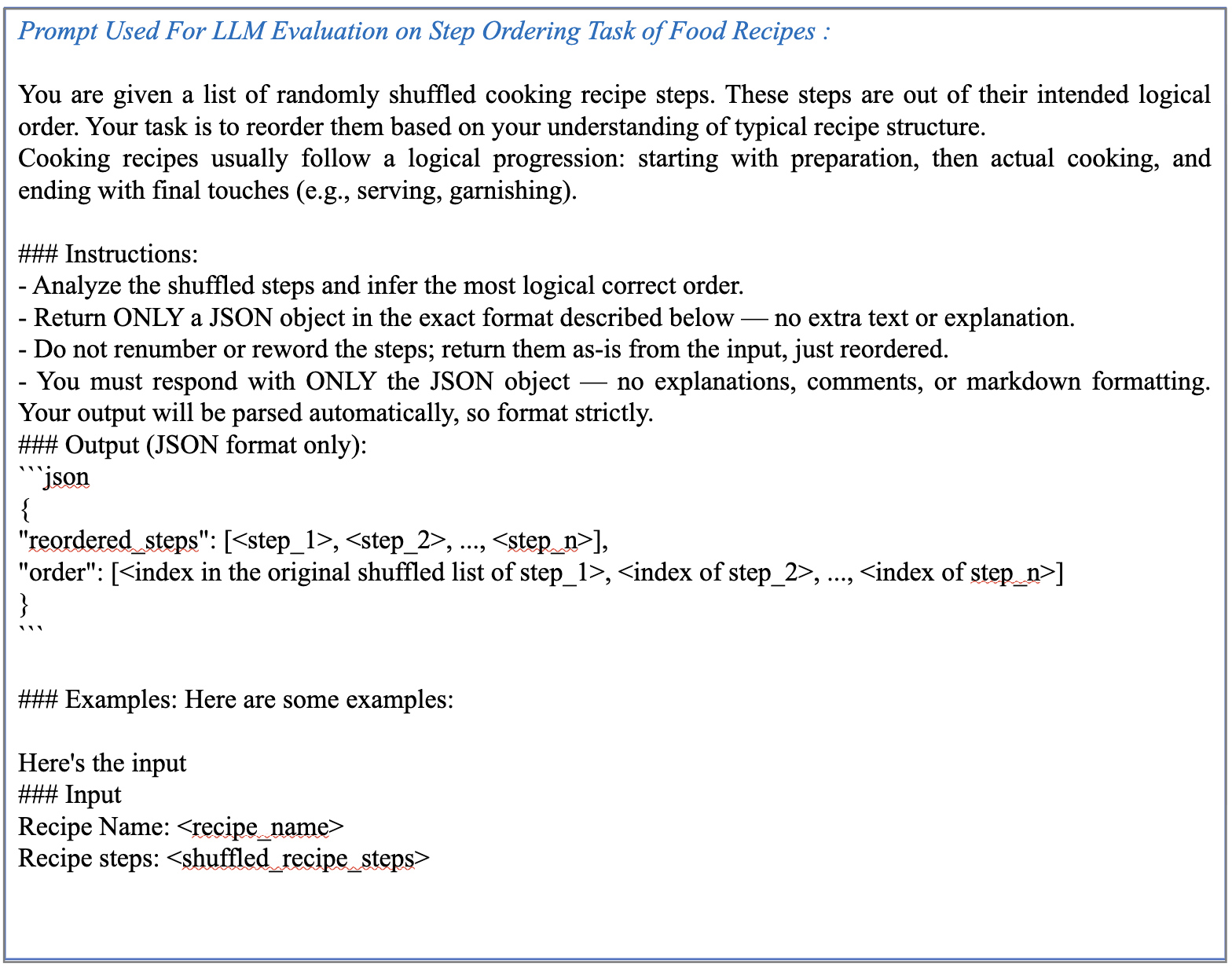}
\caption{LLM prompt used for step ordering}
\label{pt2}
\end{figure}

\clearpage

\section{Experimental Details}
\label{sec:appendix C}
The table shows the hyperparameters of the LLM models used for experimentation and their respective values. We used 1 A100 GPU for all experiments. For Qwen3-8B, the maximum number of generated tokens was set to 2048, and thinking mode was enabled, whereas for all other models `max\_new\_tokens` was 512.

\begin{table}[h!]
\centering
\begin{tabular}{|c|c|}
\hline
\textbf{Hyperparameter} & \textbf{Value} \\
\hline
temperature & 0.9 \\
\hline
max\_new\_tokens & 512 (2048 for Qwen3-8B) \\
\hline
top\_p & 0.9 \\
\hline
\end{tabular}
\caption{Hyperparameter Values}
\end{table}

\begin{table}[h]
\centering
\begin{tabular}{|c|c|c|}
\hline
\textbf{Model} & \textbf{Details} & \textbf{License} \\
\hline
LLaMA-3.1 & meta-llama/Llama-3.1-8B (Hugging Face) & llama 3.1 \\
\hline
Mistral-7 & mistralai/Mistral-7B-Instruct-v0.2 (Hugging Face) & apache-2.0 \\
\hline
Gemma-2 & google/gemma-2-9b-it (Hugging Face) & gemma \\
\hline
GPT-4o & gpt-4o-mini (OpenAI) & proprietary \\
\hline
Qwen-3 & qwen-3-8b (Hugging Face) & apache-2.0
\\
\hline
\end{tabular}
\caption{List of models used in our experiments.}
\label{tab:models}
\end{table}

\section{AI Assistance}
We have used ChatGPT for writing assistance in
the paper writing

\end{document}